# FEATURE SELECTION USING FISHER'S RATIO TECHNIQUE FOR AUTOMATIC SPEECH RECOGNITION

Sarika Hegde[1], K. K. Achary[2] and Surendra Shetty[3]

[1]Department of Computer Applications, NMAM.I.T., Nitte, Karkala Taluk, Udupi, Karnataka, India
[2]Yenepoya Research Centre, Yenepoya University, Mangalore, India
[3]Department of Computer Applications, NMAM.I.T., Nitte, Karkala Taluk, Udupi, Karnataka, India

## ABSTRACT

*Automatic Speech Recognition (ASR) involves mainly two steps; feature extraction and classification (pattern recognition). Mel Frequency Cepstral Coefficient (MFCC) is used as one of the prominent feature extraction techniques in ASR. Usually, the set of all 12 MFCC coefficients is used as the feature vector in the classification step. But the question is whether the same or improved classification accuracy can be achieved by using a subset of 12 MFCC as feature vector. In this paper, Fisher's ratio technique is used for selecting a subset of 12 MFCC coefficients that contribute more in discriminating a pattern. The selected coefficients are used in classification with Hidden Markov Model (HMM) algorithm. The classification accuracies that we get by using 12 coefficients and by using the selected coefficients are compared.*

## 1. INTRODUCTION

Feature extraction is one of the important steps in any pattern recognition task. The efficient feature extraction technique extracts the feature which is able to discriminate one pattern from another accurately. There are many feature extraction techniques available for representing speech sound pattern in Automatic Speech Recognition (ASR) task. Mel Frequency Cepstral Coefficients (MFCC) and Linear Predictive Coefficients (LPC) have been identified as the most successful feature representation for the speech signal [1]. But recently, MFCC gained more importance than LPC due to its high discriminating capability [2]. MFCC is one of the signal processing techniques when applied on speech sound, extract the 12 coefficients. This set of 12 MFCC coefficients is used as a feature vector in classification task.

If the feature vector is of high dimension, it is reduced to lower dimensional subspace using the techniques like Principal Component Analysis (PCA) and Linear Discriminant Analysis (LDA). These techniques are also known as dimensionality reduction techniques. Dimensionality reduction of the features is done, so that the computational cost and system complexity for subsequent processing can be decreased. Linear Discriminant Analysis [3][4][5] finds a linear transform by maximizing the Fisher's ratio and is found to be good in discriminating patterns.

PCA technique computes the feature vector subspace with maximum variance [6]. These techniques are also used successfully in classification with HMM [7][8][9].

Usually the above techniques are applied when the dimension is very large and when the features are computed using more than one feature extraction techniques. But, while using MFCC features for ASR, usually all the 12 coefficients are considered in the feature vector. There is no clear guideline in literature that whether the subset of 12 coefficients can be used as feature vector instead of using all the 12 MFCC. Rarely, one can find a work where, a specific technique is used to select the subset of 12 MFCC coefficients for classification. One can find that the Fishers ratio (F-ratio) technique is used in Linear Discriminant Analysis (LDA) as computing the linear combination of feature vectors which maximizes the F-ratio value. It is also sometimes used to compare the two feature extraction technique for their capability of discriminating patterns in classification [2][10][11]. F-ratio technique is used in this work for selecting the subset of MFCC coefficients best suitable for classification.

The recorded audio sounds of *five* vowels in *Kannada* language is considered as dataset. The objective is to apply MFCC feature extraction technique on this dataset and classify using Hidden Markov Model (HMM) classification algorithm. It is analyzed, whether the 12 coefficients can be replaced by a subset of coefficients as feature vector for classification without degrading the classification accuracy. Using Fisher's ratio (F-ratio) as the criterion for selecting the subset of MFCC coefficients, computational experiment has been carried out for classification of vowels. F- ratio uses between vowel class variance and the within vowel class variance for each coefficients and thus reflects the capability of an MFCC coefficient in discriminating one vowel class with others.

The organization of the rest of the paper is as follows. The second section describes MFCC feature extraction technique, Fisher's ratio technique and HMM classification technique. In the third section, the results of the experiments are discussed and analyzed. The last section concludes the work and highlights the future work also.

## 2. METHODOLOGY

In this section, the techniques used for feature extraction, F-ratio calculation and classification are explained.

### 2.1. Mel Frequency Cepstral Coefficient

The Mel Frequency Cepstral Coefficient (MFCC) is a feature extracted by applying more than one Fourier Transform sequentially to the original signal [6][12]. The first step is preprocessing which consists of framing and windowing of the signal. Framing is the process of breaking the set of sample observations of an entire audio file into smaller chunks called as frame [13]. We have used a frame size of 30ms which is normally used in speech recognition applications. For each of the frame $f_i$ of size N, DFT coefficients are calculated by applying the following equation.

$$X_i[m] = \sum_{n=0}^{N} x[n]e^{imn2\pi/N}, 1 \le m \le M \qquad (1)$$

The resulting value $X_i[m]$ is a complex number and the power spectrum for this is computed as,

$$P_i[m] = \frac{1}{N}|X_i[m]|^2, \qquad (2)$$

The power spectrum is then transformed to mel frequency scale by using a filter bank consisting of triangular filters, spaced uniformly on the mel scale. To start with, the lower and higher values of frequencies are converted into mel scale. An approximate conversion between a frequency value in Hertz ($f$) and mel is given by:

$$mel(f) = 2595 \log_{10}\left(1 + \frac{f}{700}\right) \qquad (3)$$

Based on the number of filters to be designed, the intermediate mel frequencies are computed. Each of the mel frequency is then converted back into normal frequency with the inversion formula. By doing this we get the list of frequencies with mel-scale spacing. For each of these frequency responses, a triangular filter is created and the series of such filters is called as mel filter bank. Each triangular filter is then multiplied with power spectrum $P_i[m]$ and the coefficients are added up, which will be an indication of how much energy is within each filter. Finally, the cepstral coefficients are calculated from these calculated energy values of filter-bank by applying Discrete Cosine Transform (DCT) of the logarithm of the filter-bank energy. This is given by,

$$C_i = \sum_{k=1}^{K} (\log S_k).\cos\left(\frac{i\pi}{K}\left(k - \frac{1}{2}\right)\right) \quad k = 1,2,\ldots,K \ \& \ i = 1,2\ldots,L \qquad (4)$$

'L' is the number of MFCC coefficients considered, $C_i$ is the $i^{th}$ MFCC coefficient, $S_k$ is the energy of $k^{th}$ filterbank channel (i.e. the sum of the power spectrum bins on that channel), 'K' is the number of filterbanks. A more detailed description of the mel-frequency cepstral coefficients can be found in [6].

## 2.2. Fisher's Ratio (F-ratio) Technique

A set of MFCC feature measurements would be effective in discriminating between vowels if the distributions of different vowels are concentrated at widely different locations in the parameter space. For a feature value, a good measure of effectiveness would be the ratio of inter vowel to intra-vowel (within class) variance, often referred to as the F-ratio [10][14]. Consider the MFCC feature dataset for the *five* vowels given as,

$$X = (x_{i1}, x_{i2}, \cdots, x_{i12})$$

where the value of 'i' varies from 1 to 'L' number of observations. F-ratio is given as,

$$F = \frac{\text{Between vowel variance for an MFCC coeffecient}}{\text{Average of within vowel variance for an MFCC coeffecient}}$$

It can also be represented as,

$$F = \frac{\frac{1}{N}\sum_{i=1}^{N}(\mu_i - \bar{\mu})^2}{\frac{1}{N}\sum_{i=1}^{N} S_i} \qquad (5)$$

N, indicates the number of vowels, $\mu_i$ is the mean of MFCC coefficient for $i^{th}$ vowel class, $\overline{\mu}$ is the overall mean value for the MFCC coefficient from all the vowel classes. $S_i$, the within vowel variance is given by,

$$S_i = \frac{1}{M_i} \sum_{j=1}^{M_i} (x_{ij} - \mu_i)^2$$

(6)

$x_{ij}$ is the MFCC coefficient for $j^{th}$ observation of $i^{th}$ vowel and $M_i$ is the number of observations in $i^{th}$ vowel . This technique is applied for all the 12 MFCC coefficients individually. Higher F-ratio value for an MFCC coefficient indicates that it can be used for good classification [10][11].

## 2.3. Hidden Markov Model (HMM)

A Hidden Markov Model (HMM) is a statistical model in which the system being modelled is assumed to be a Markov chain with unobserved (*hidden*) states [6]. A Markov chain is a stochastic process in which the future state of the process is based solely on its present state. HMM is a very common technique in classification, especially in the case of sequential data processes such as speech, music and text. HMMs are used to specify a joint probability distribution over hidden state sequences S={s₁, s₂,..., s_T} and observed output sequences X={x₁, x₂,..., x_T} . The logarithm of this joint distribution is given by:

$$logP(X,S) = \sum [\log P(s_t|s_{t-1}) + logP(x_t|s_t)]$$

(7)

Here the hidden states $s_t$ and observed outputs $x_t$ denote vowels labels and corresponding feature vectors respectively. The distributions P(x_t|s_t) are typically modelled by multivariate Gaussian mixture models (GMMs).

Where, N(x_t; μ, Σ) denotes the Gaussian distribution with mean vector μ and covariance matrix Σ, and M denotes the number of mixture components per GMM. The mixture weights $w_{jm}$ are constrained to be nonnegative and normalized:

$$\sum_{m=1}^{M} w_{jm} = 1$$

(8)

Let $\theta$ denote the vector of model parameters including transition probabilities, mixture weights, mean vectors, and covariance matrices. The goal of parameter estimation in HMMs is to compute the optimal $\theta *$ given N pairs of observations and target label sequences {X_i, Y_i} where Y_i is the class label [15][16].

## 3. RESULTS AND DISCUSSIONS

For the experiment, audio sounds of 5 vowels in *Kannada* language are used. The list of vowels is shown in the following table (Table 1) with their corresponding Unicode name. *Twenty five* patterns of each vowel are recorded with the voice of single female *Kannada* speaker constituting totally 125 audio clips.

Table 1. List of Vowels

| Vowels |
| --- |
| ಅ (/a/) |
| ಇ (/i/) |
| ಉ (/u/) |
| ಎ (/e/) |
| ಒ (/o/) |

Each of the audio sound is divided into frames with duration of 30ms. Then the frames are processed to extract 12 MFCC coefficients called as a feature vector. For each audio file, we get number of feature vectors equal to the number of frames. The entire dataset of MFCC coefficients are analyzed with Fisher's ratio technique. F-ratio is computed for each coefficient individually using the steps explained in section 2.2. The result of this computation is shown in Figure 1, where x-axis indicates the MFCC coefficient number and y-axis indicates the corresponding F-ratio value.

From the figure, it can be observed that for the 3$^{rd}$ MFCC coefficient, F-ratio value is highest. Does it mean that the 3$^{rd}$ MFCC coefficient for vowel sound has high discriminating capability than other coefficient? To analyze this property, we have retrieved a subset of MFCC coefficients with high F-ratio and used this subset as a feature vector in classification. The classification experiments are done by varying the number of MFCC coefficients for subset selection. The number of coefficients considered is varied from 3 to 12 and is selected in the order of high F-ratio value. The classification is done using HMM algorithm.

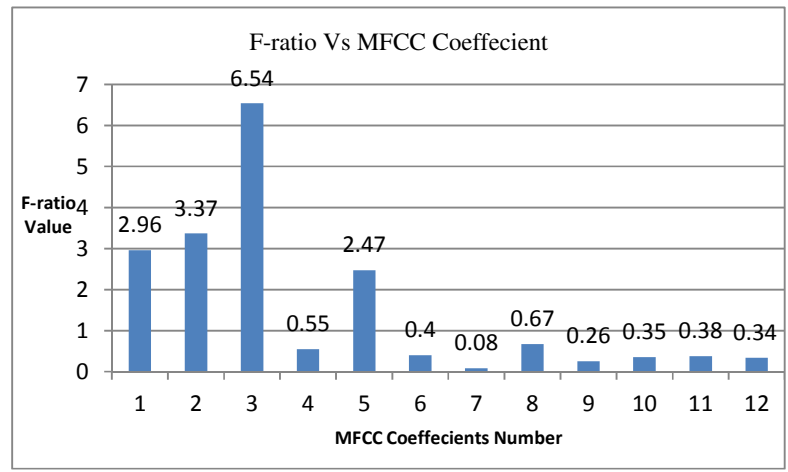

Figure 1        F-Ratio value for 12 MFCC coefficients

In the process of classification, first the dataset is divided into training and testing set using hold out approach. Using the training datasets, classification parameters like, prior probability table, emission table and transition table are computed for each of the vowel class separately. Here we get five HMM models corresponding to each vowel class. In the process of testing, the test data is tested against each of the HMM models individually to compute the maximum likelihood probability. The HMM model against which the test data has highest likelihood probability is considered as the predicted output. If the predicted output is matching with the actual output we consider it as correct prediction. In this way, accuracy of classification is computed. More detail on how HMM is used for classification can be found in [17]. Since the dataset is randomly divided into testing and training set, each time we repeat the experiment we get different training and testing dataset. We have repeated 50 iterations of classification for each dataset and found the result getting saturated over 50 iterations. The average accuracy over fifty iterations is considered as the final classification accuracy. The following table (Table 2) summarizes the classification accuracy result for different types of experiment.

The following observations can be made from Table 2. The number of coefficients selected has an effect on the classification accuracy. The classification accuracy increases as the number of coefficients increases but only up to certain point. Selection of 8 MFCC coefficients gives a classification accuracy which is comparable or even better than selecting all 12 coefficients. This clearly shows that there is an effect of selecting the MFCC coefficients based on high F-ratio value. Instead of using all the MFCC coefficients as feature vector for classification, we can use the 8 coefficients selected in the order of high F-ratio values. This technique also helps in reducing the feature dimension without much affecting the classification accuracy.

Table 2. Classification accuracy

| No. of MFCC Coefficients | Accuracy (%) |
|---|---|
| 3 MFCC | 70.07 |
| 4 MFCC | 70.83 |
| 5 MFCC | 72.29 |
| 6 MFCC | 72.25 |
| 7 MFCC | 74.94 |
| 8 MFCC | **76.45** |
| 9 MFCC | 75.41 |
| 10 MFCC | 74.83 |
| 11 MFCC | 75.54 |
| 12 MFCC | **74.72** |

## 3. CONCLUSIONS

Computational experiments are done to analyze whether the number of MFCC coefficients used in feature vector can be reduced using Fisher's ratio measure. The experimental results show that there is improvement in classification accuracy when eight MFC coefficients are selected based on F-ratio criterion compared to the conventional method of using twelve MFC coefficients. The classification accuracy is affected based on the number of MFCC coefficient selected. F-ratio technique has been proved promising in the selection of subset of MFCC feature. We are conducting more experiments for analyzing whether the same effect is found for consonants and simple syllables.

## AUTHORS

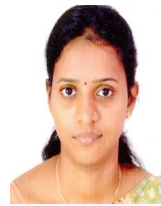

**Sarika Hegde** received B.E (Information Science) degree from NMAM.I.T Nitte in 2003 and M.Tech (Computer Science) from NMAM.I.T Nitte in 2007. Presently she is working as Associate Professor in the department of Computer Applications at NMAM.I.T Nitte and also pursuing PhD in the area of Automatic Speech Recognition at Mangalore University, Mangalore. Her area of interest includes Speech Recognition, Audio Data Mining, Big Data, and Pattern Recognition.

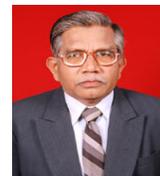

**Dr. K. K. Achary** is a Professor of Statistics & Biostatistics in Yenepoya Research Centre, Yenepoya University, Mangalore, India. He holds M.Sc degree in Statistics and PhD in Applied Mathematics from Indian Institute of Science, Bangalore, India. His current research interests include Stochastic Models, Inventory Theory, Face Recognition Techniques, Audio Data Mining and Speech Recognition. His research papers have appeared in European Journal of Operational Research, Journal of Operational Research Society, Opsearch, CCERO, International Journal Information and Management Sciences, Statistical Methods and American Journal of Mathematics and Managem ent, International Journal of Speech Technology.

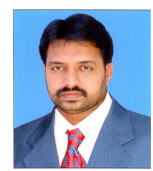

**Dr. Surendra Shetty** is working as Professor in the Department of Computer Applications, NMAM.I.T, Nitte, Udupi, India. He holds MCA, MBA (HR) degrees. He did PhD in the area of Au dio Data Mining. His main research area includes Audio Data Mining, Speech Recognition, Big Data, Data Mining, Management Information System, and Software Testing. He has presented and published more than 10 papers in Journals and conferences.